\documentclass[runningheads]{llncs}
\usepackage[T1]{fontenc}
\usepackage{graphicx}
\usepackage{multicol}
\usepackage{pslatex}
\usepackage{algorithm2e}
\usepackage[bottom]{footmisc}
\usepackage{multirow}
\usepackage{bbding}
\usepackage{amsmath}
\usepackage{hyperref}

\begin{document}
\title{Scene-aware Prediction of Diverse Human Movement Goals}
%
%
\author{Qiaoyue Yang\inst{1}\orcidID{0009-0009-8529-0652} \and
Amadeus Weber\inst{1}\orcidID{0009-0009-7335-8488} \and
Magnus Jung\inst{2}\orcidID{0009-0006-1353-4185} \and
Ayoub AI-Hamadi\inst{2}\orcidID{0000-0002-3632-2402} \and
Sven Wachsmuth\inst{1}\orcidID{0000-0001-5371-7214}}
\authorrunning{Q. Yang et al.}
%
\institute{COSY@Home-Lab, Faculty of Technology, Bielefeld University, 33619 Bielefeld, Germany\\  \email{\{qyang, amweber, swachsmuth\}@techfak.uni-bielefeld.de}\and
Institute for Information Technology and Communications (IIKT), Otto von Guericke University Magdeburg, 39106 Magdeburg, Germany\\
\email{\{magnus.jung, ayoub.al-hamadi\}@ovgu.de}}
\maketitle              
\begin{abstract}
Anticipation of human behaviours facilitates autonomous systems in proactive planning. Human behaviour could be stochastic due to varying goals. Human goals typically guide their own movement and could therefore help to predict the human trajectory and human motion in the long-term. To infer the human movement intentions, the environmental context plays a significant role, in addition to the social cues expressed by the individual. Previous works on human goals prediction either require semantic knowledge of the scene, or only tackle interactions with objects. In this paper, we propose a novel multi-goal prediction method using the generative model to address the stochasticity of human movement. It leverages the current RGB scene and the human pose to predict diverse potential future goals of human movement based on the Conditional Variational Autoencoder (CVAE). Our results demonstrate that our approach is capable of generating multiple movement goals in the scene via samplings in latent space of the CVAE and exhibits generalization capability across scenarios in GTA-IM dataset and PROX dataset. Code is publicly available at \href{https://github.com/Q-Y-Yang/DiverseGoalsPrediction.git}{\texttt{https://github.com/Q-Y-Yang/DiverseGoalsPrediction}}.

\keywords{Human Intentions; Human Goals Prediction; Probablistic Motion Prediction; Variational Autoencoder.}
\end{abstract}
\section{Introduction}
Human behaviour is inherently stochastic, for example, a person could suddenly change his or her mind while being engaged in an activity. But this is often driven by goals. To understand human movement intentions, human trajectory prediction resolves 2D waypoints along the path and human motion/pose prediction tackles 3D full body motion. In order to cover uncertainties in the future, probabilistic forecasting attracts more attention than deterministic methods \cite{review,survey}.  Long-term human trajectory can be planned conditioned on the future movement destinations \cite{endpoint,goalswaypoints,socialtraj,goaldriven}, and these waypoints also contribute to the accuracy of long-term human motion predictions \cite{longterm,motion2}. The intention of a person is often implicitly hidden in the mind because people do not always express them verbally. But it is usually associated with the surrounding environment, for example, the desire to interact with a specific object such as a table, chair, cup, coffee machine, fridge, etc. \cite{navigoal}, or to navigate in the environment such as passing through a doorway and walking up the stairs.

In human robot cooperation, it is imperative for robots to possess visual perception capabilities towards humans as partners, rather than merely detecting humans as common obstacles. For instance, these applies to environments where humans and robots coexist, such as public areas like museums and shops where robots provide services to customers \cite{hrcbook}, and home scenarios where robots perform as a helper to healthcare \cite{hrcbook} and housework. In robot navigation, a multi-layer costmap \cite{navigation} is usually deployed to guide the global and local planning. Among them, a social navigation layer \cite{sociallayer} provides the possibility to adjust the costs for navigating around humans. Predicting potential goals of human partners benefits the robot's behavioral planning in a proactive and safe way, for instance, avoiding impeding the potential movement areas of human counterparts in advance \cite{navigoal}. Besides, in autonomous driving, self-driving cars should also be capable of anticipating pedestrians' potential movement destinations, in order to react in advance to avoid accidents. 

\begin{figure}[h]
	\begin{tabular}{cccc}
		\includegraphics[width=0.245\textwidth]{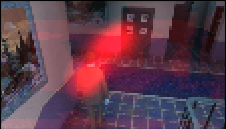}&
		\includegraphics[width=0.245\textwidth]{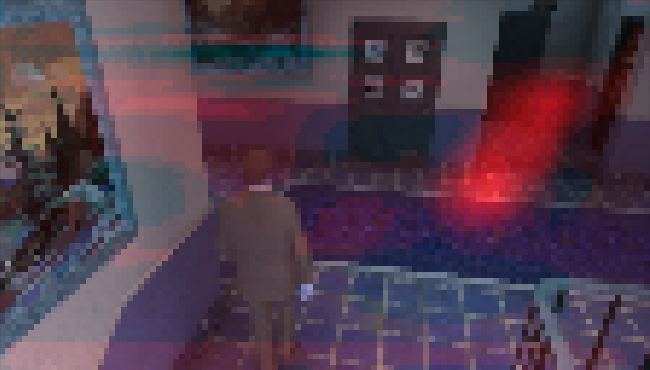}&
		\includegraphics[width=0.245\textwidth]{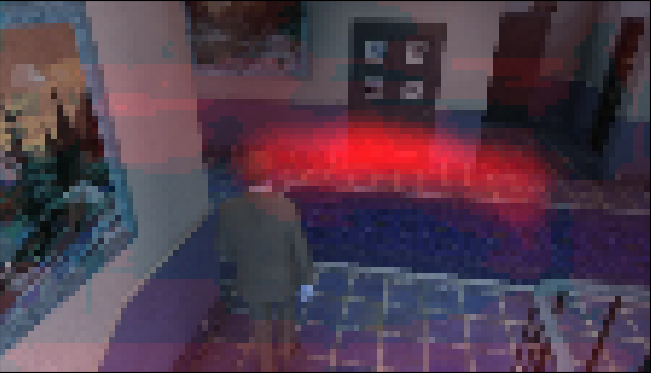}&
		\includegraphics[width=0.245\textwidth]{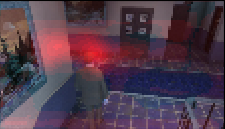}\\
		(a)&(b)&(c)&(d)
	\end{tabular}
	\caption{\textbf{Predictions of possible movement areas in the future.} (a) The subject walks straight to the door. (b) The subject turns right towards the doors. (c) The subject turns right into the corridor. (d) The subject turns left into the corridor.}
	\label{fig1}
\end{figure}

Possible human goals in the scenario could help overcome the difficulty of long-term forecasting of human trajectories and human motions by serving as priors. These future goals could also directly benefit proactive autonomous systems. In this work, we propose a method based on Conditional Variational Autoencoders (CVAE) that forecasts scene-aware future goals of the human movement from only a single frame of the current scene and the human pose. Figure \ref{fig1} demonstrates scene-compliant predictions of the human movement goals. The contributions of this paper can be summarized as: 

(1) An end-to-end method for predicting scene-compliant targets of long-term human motion, based on the current RGB scene image and the human pose.

(2) Addressing stochastic nature of human behaviour by anticipating multiple possible targets, without the need for multiple goal positions as labels.

\section{\uppercase{Related Work}}

\paragraph{Goal-directed Forecasting.} Previous works leverage goal prediction to address long-term human trajectory and motion prediction problems \cite{samedata1,samedata2,endpoint,goalswaypoints,socialtraj,longterm,motion2,goaldriven}. \cite{samedata1,samedata2} forecast deterministic human motion along a set of contact points in the scene point cloud, but they do not take into account multiple potential human intentions. \cite{longterm} predict human motion goals based on a sequence of heatmaps of  past human joints and a scene image to benefit long term 3D motion prediction. However, a set of joint heatmaps during a past time slot might not fit in a single scene frame because the scene may have changed over time. Additionally, generating a sequence of heatmaps of human joints is not computational efficient during inference on the real robot, because the time required increases linearly with the number of heatmaps. In the real application, the requirement for a fixed length of history reduces flexibility, i.e. the prediction system can only work if such a length of history is available. \cite{graph} also utilizes a series of observed 2D human poses and images to forecast 3D human poses. However, it only predicts a fixed future path and poses, neglecting the inherent stochasticity of human behaviors.

In outdoor scenarios from bird view, a semantic map is often required to incorporate scene information and derive movement goals. For instance, Y-net \cite{goalswaypoints} uses a semantic segmentation map of the scene and the past trajectory to forecast goals and waypoints based on U-Net architectures. The SNS-LSTM model \cite{socialtraj} takes the observed trajectory, a navigation map and a semantic map as inputs to extract future positions by LSTM (long short-term memory). GoalGAN \cite{tumgoal} also firstly predicts goals via a convolutional neural network (CNN) from an RGB scene and  dynamic features of the pedestrian. PECNet \cite{endpoint} proposes an endpoint variational autoencoder (VAE) to predict final goals of all pedestrians based on history trajectories and ground truth endpoints, but the ground truth endpoints are not available during inference. \cite{goaldriven} divides the boundary of the scene into grids as potential destinations. However, other areas of the scene are neglected and dividing into grids could corrupt the scene information by introducing artifacts.

In home scenarios, the layout of the environment could be changed more frequently, compared to public environments such as streets, museums, and train stations. For instance, tables and chairs at home can be easily moved. If the forecasting relies on a pre-established model of the environment, then the home model has to be frequently adjusted.

In addition, some approaches concentrate on human object interactions. The probabilistic transition model \cite{navigoal} predicts the human's navigation goal in accordance with past human poses and object interactions. \cite{motion2} propose a GoalNet that estimates positions and directions of contact points between the human and the interactive object with known object geometry. Nevertheless, they can only handle targets related to interactive objects, and do not cover other aspects of an entire scene.

\paragraph{Variational Autoencoders (VAEs).} VAE as one of the generative models has achieved promising performances in various applications such as image reconstruction, image generation, representation learning, semi-supervised learning, anomaly detection, and data augmentation \cite{vaereview}. An Autoencoder (AE) learns a compressed representation of the input data via an encoder network and reconstructs the input from that compressed latent space through a decoder network \cite{vaecompare}. VAE \cite{vae} further extends the compressed space to a probabilistic distribution such as a Gaussian distribution and generates new samples from that distribution. The Conditional Variational Autoencoder (CVAE) \cite{cvae} was originally proposed to generate new handwritten digits from the MNIST dataset by conditioning on 10 classes of digits. It further proved great performances in face image manipulation conditioned on a set of face attributes such as gender, expression, hair colour, age, and eyewear etc. \cite{cvaeface}.

\section{\uppercase{Proposed Method}}

\subsection{Problem Formulation}
The problem we address in this work is to predict human movement goals $G$ in the future time $T$ from the perspective of the current time $0$.  To cover the stochastic nature of human behaviour, multiple goals $G^{F}_T$ should be resolved, given the human $H_0$ and the scene $S_0$ at zero time, as in Equation \ref{eq1}.  The future time horizon is defined by a criterion that the future position at time $T$ is already present in the scene at time $0$.
\begin{equation}
	\left(S_0, H_0\right) \rightarrow G^{F}_T.
	\label{eq1}
\end{equation}

\subsection{Proposed Approach} 
Our approach forecasts interactive goals by learning features of goal areas, such as walking along a corridor, going towards a chair, instead of relying on a past trajectory as previous methods did. Given a past trajectory, it is possible to calculate the speed and direction of movement, and therefore estimate the future position. However, this only works if a fixed-length history is available and if the person does not change their mind during this time period. Instead, our approach predicts potential interaction goals in the scene at every moment by sampling in a 200-dimensional latent space.

\begin{figure*}
	\begin{center}
		\includegraphics[width=\textwidth]{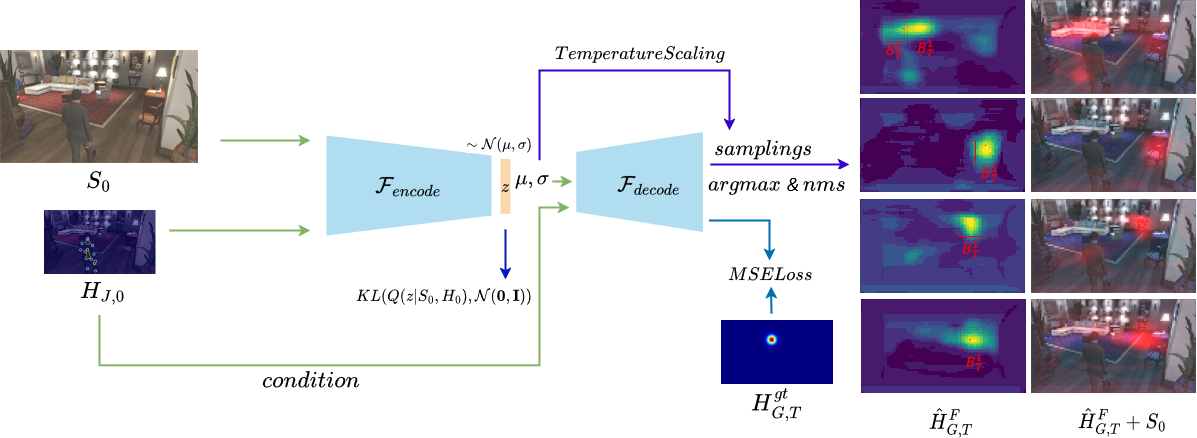}
	\end{center}
	\caption{\textbf{Method.} The encoder takes a scene image and a heatmap of human joints as input. The encoded latent vector $z$ and the heatmap of human joints as condition are fed to the decoder. In the training stage, the MSE loss between the ground truth goal heatmap and the decoder output, and the KL divergence between the posterior of $z$ and the Gaussian distribution are computed to optimise the weights of the CVAE (blue arrows only exist during training). During inference, temperature scaling is used to generate multiple samples as diverse goal areas (purple arrows only work during inference). The results $\hat{H}_{G, T}^{F}$ show the predicted destination in the sofa, the chair, the desk and the way to another room respectively (from top to bottom).}
	\label{pipeline}
\end{figure*}

Figure \ref{pipeline} illustrates the overall pipeline of our approach. The model $\mathcal{F}$ takes a scene image $S_0$ and a heatmap of human joints $H_{J, 0}$ at current time $0$ as inputs, to anticipate multiple goals $G^{F}_T$ with corresponding confidence scores $C^{F}_T$ at frame $T$ in the future,
\begin{equation}
	\left(G^{F}_T, C^{F}_T\right) = \mathcal{F} \left(S_0, H_{J, 0}\right),
\end{equation}
where $F$ refers to the number of forecasts, and $T$ is defined to $2$ seconds in our case.

Concretely, the encoder $\mathcal{F}_{encode}$ learns a latent vector $z$ from the inputs, and the decoder $\mathcal{F}_{decode}$ predicts goal heatmaps $\hat{H}_{G, T}$ at frame $T$ from the samplings of the latent vector $z$ with the condition of the joint heatmap $H_{J, 0}$,
\begin{equation}
	z \sim \mathcal{N}(0, \mathbf{I}) = \mathcal{F}_{encode} \left(S_0, H_{J, 0}\right),
\end{equation}
\begin{equation}
	\hat{H}_{G, T} = \mathcal{F}_{decode} \left(H_{J, 0}, z\right).
\end{equation}

The assumption of the Gaussian distribution for the approximated posterior $q_{encode}(\mathbf{z} \mid \left(S_0, H_{J, 0}\right))$  makes gradient decent more efficient \cite{vaereview}.  The distribution of the latent variable $p(\mathbf{z})$ is also assumed to be a high-dimensional mixture of Gaussians with zero mean and unit variance $\mathcal{N}(0, \mathbf{I})$ \cite{tutorial}. The dimension of our latent vector $\mathbf{z}$ is set to $200$, since higher dimensions have stronger capability to represent complex features of scenes \cite{generativebook,sampling}. Lighter areas of the output $\hat{H}_{G, T}$ indicate higher probabilities of being goal areas.

%


\subsection{Training and Loss Function.} 
We use the torso position at time $T$ as the ground truth goal position at time $0$ and generate the corresponding ground truth goal heatmap $H_{G, T}^{gt}$, so that manual labelling is not necessary. In particular, our method does not require labelling of multiple goals within the scene, despite the capability to predict various goals.

The model is trained with a loss function $L$ consisting of a KL-divergence $L_{KL}$ between the variational posterior $q_{encode}(\mathbf{z} \mid \left(S_0, H_{J, 0}\right))$ of encoder and a normal distribution $\mathcal{N}(\mathbf{\mu}, \sigma \mathbf{I})$, and Mean Squared Error (MSE) $L_{MSE}$ between the predicted goal heatmap $\hat{H}_{G, T}$ and the ground truth goal heatmap $H_{G, T}^{gt}$,
\begin{align}
	L &= L_{KL} + L_{MSE},\\
	L_{KL} &= KL[q_{encode}(\mathbf{z} \mid \left(S_0, H_{J, 0}\right)) || \mathcal{N}(\mathbf{\mu}, \sigma \mathbf{I})],\\
	L_{MSE} &= MSE(\hat{H}_{G, T}, H_{G, T}^{gt}).
\end{align}

Unlike other goal prediction methods \cite{longterm,tumgoal}, a $L_2$ loss that measures distance between the predicted goal position and the ground truth goal position is not chosen in our approach. Our ground truth is not just a coordinate of the target position, but a heatmap generated by a Gaussian distribution centred on the target coordinate. The size of the generated goal area is determined by the criterion that the features of the target area are covered. A too large area of the ground truth goal results in greater inaccuracy of goal positions and a too small area lacks features. The purpose of using $MSE$ loss between the predicted and ground truth heatmaps is to implicitly learn features of goal areas from the scene, which benefits the generalization of predictions. Comparably, \cite{tumgoal} uses a cross entropy loss for the probability that a predicted grid cell corresponds to the ground-truth goal. However, the grid segmentation of the scene could corrupt spatial information, for instance, a door may be divided into two or more grid cells.

\subsection{Inference via Generation}\label{generate}
We generate more than one prediction $\hat{H}_{G, T}^{(f)}$ via temperature scaling (changing the variance $\sigma$ of the latent vector $z$) \cite{temperature} to address the stochasticity of human behaviour during inference, even though the ground truth goal is only fixed during training. Equation \ref{temperature} describes the reparametrization trick \cite{vae} of VAE with temperature scaling:
\begin{equation}
	z = \mu+\sigma \cdot \varepsilon \cdot \tau,
	\label{temperature}
\end{equation}

where $ \varepsilon \sim N(0,1)$ works as white noise, $\tau$ refers to the temperature that controls diversity of generation. A higher scaling $\tau$ leads to more variety to cover potential possibilities for future goals. It turns to deterministic prediction when $\tau \rightarrow 1$, which can be applied to scenes such as stair climbing where there are not many behavioural options.

In the post-processing stage, bounding boxes of possible goal areas $B^{N}_T$ are extracted by selecting $N$ $argmax$ of each generated goal heatmap $\hat{H}^{(f)}_{G, T}$ (Equation \ref{tab2}). These raw bounding boxes $B^{N}_T$ are then merged by performing non-maximal suppression (NMS) \cite{nms} to propose $M$ bounding boxes predictions $B^{M}_T$ with corresponding confidence scores $C^{M}_T$ (Equation \ref{tab3}). The purpose of this is to take into account local maxima rather than solely extracting one $argmax$, as the output heatmap often displays multiple local maxima. Finally, each predicted goal $G^{(m)}_T$ is extracted by $argmax$ operation from every proposed bounding box $B_T^{(m)}$ (Equation \ref{tab4}). The goals $G^M_T$ can be ranked by their corresponding confidence scores $C^M_T$, and the goal with the highest confidence score can be selected as a final prediction. In particular, the size of the bounding boxes is determined by the size of the ground truth target area, the former being slightly larger than the latter to ensure that the bounding boxes are able to detect the region of interest.

\begin{align}
	B^{N}_T = argmax(\hat{H}^{(f)}_{G,T}),\label{tab2}\\
	B^{M}_T, C^{M}_T = NMS (B^{N}_T), \label{tab3}\\
	G^{M}_T = argmax(B^{M}_T),\label{tab4}
\end{align}
where $N$ denotes the number of raw bounding boxes, and $M$ denotes the number of proposed bounding boxes with the corresponding confidence scores.

\section{\uppercase{Experiments}}
\subsection{Experimental Setups}
\paragraph{Datasets.} GTA Indoor Motion dataset (GTA-IM) \cite{longterm} is a synthetic dataset that contains various human behaviors such as going upstairs and downstairs, walking towards another room, going to sit on a chair or sofa, lying on a bed, going to drink in the kitchen, etc. It also includes 10 different scenes, multiple subjects, and provides 21 human joints positions in the camera coordinate and in the world coordinate. Notably, the camera in the GTA game engine moves along with the subject so that it keeps capturing the human, which is similar with cameras on mobile robots rather than fixed cameras in the space. PROX (Proximal Relationships with Object eXclusion) \cite{prox} is a real-world dataset that records human movements in and interactions with the world. It includes 12 scenarios such as bedrooms, living rooms, offices, and sitting booths \cite{prox}. 10 of these were used for training, and the remaining two scenarios were used to test the model's ability to generalise across scenes. Compared to the synthetic GTA-IM dataset \cite{longterm}, the scenes of PROX data are more compact and disordered. For instance,  more pieces of furniture are placed in the room, facing in different directions.

\paragraph{Evaluation Metric.} Final Displacement Error (FDE) \cite{FDE} measures average Euclidean distance between predicted and ground truth goals (endpoints). This metric is frequently deployed in trajectory prediction to assess the accuracy of destination forecasts, so it is also highly applicable to our research. $FDE_{avg}$, as in Equation \ref{metric}, computes the average error across multiple goal predictions for each test data, and then averages these errors over the entire test set. $FDE_{min}$, as in Equation \ref{metric_min}, selects the prediction with the smallest error for each test data and then averages these minimal errors across all test data. The assessment is performed in 2D image space in pixels. 

\begin{align}
	FDE_{avg} = \frac{1}{S} \frac{1}{F} \sum_{s=1}^{S} \sum_{f=1}^{F} \mathop{\min}_{m} \| G_{T}^{gt}-G_{T}^{(s, f, m)} \|_2,\label{metric}\\
	FDE_{min} = \frac{1}{S} \sum_{s=1}^{S} \mathop{\min}_{f} \mathop{\min}_{m} \| G_{T}^{gt}-G_{T}^{(s, f, m)} \|_2,\label{metric_min}
\end{align}
where $S$ refers to the set of test data and $F$ refers to the number of probabilistic forecasts for the test data. $ G_{T}^{gt} $ and $ G_{T}$ represent the ground truth goal coordinate and the predicted goal coordinate respectively. $m$ denotes the local maxima that yields the minimal error of every forecast $f$. For deterministic goal prediction, $F$ equals to 1, and $ F = 3, 5, 10 $ are evaluated for probabilistic prediction. 

\paragraph{Baseline.} Given the limited research on probabilistic destination forecasting in indoor scenarios, we compare our Goal CVAE with the GoalNet \cite{longterm} in GTA-IM dataset and PROX dataset. GoalNet \cite{longterm} was reproduced since it is not open-sourced, and we retrained and evaluated under the same conditions as our method. Since the baseline does not provide the information how to generate probabilistic forecasts in their paper, we deployed our Inference via Generation as explained in Section \ref{generate} on the baseline model. A 1-second history was used by the GoalNet \cite{longterm} to predict destinations 2 seconds ahead.

\paragraph{Implementation Details}
Our work is implemented in PyTorch \cite{pytorch}, trained with the ADAM optimizer \cite{adam}. 2 Epochs were trained with a batch size of 64 on NVIDIA GeForce GTX 1060. Learning rate is $10^{-4}$. In order to avoid overfitting and ensure the diversity of our predictions, the frames per second (FPS) of training data was reduced to 5, so that there are fewer duplicate inputs. 

\subsection{Results}

\subsubsection{Qualitative Analysis}
\begin{figure}[ht!]
	\centering
	\begin{tabular}{cc}
		\includegraphics[width=0.45\textwidth]{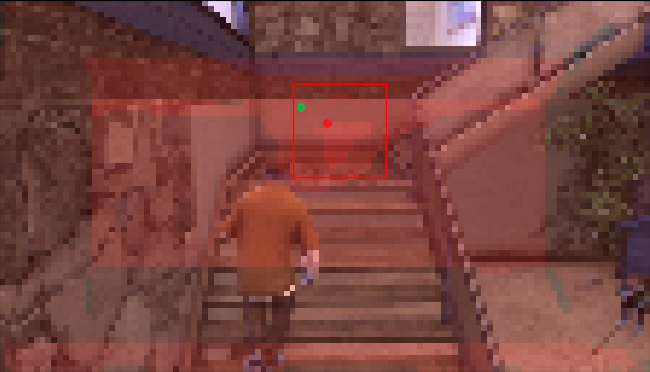}&
		\includegraphics[width=0.45\textwidth]{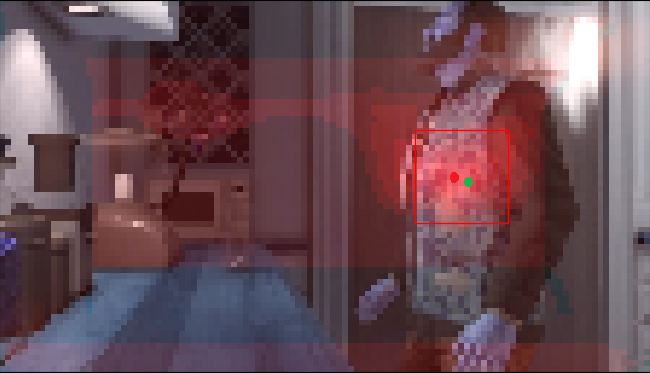}\\
		(a)&(b)
	\end{tabular}
	\caption{\textbf{Deterministic Predictions in GTA-IM.} Within the bounding boxes of the predicted goal areas, the red dots represent the predicted goal positions and the green dots represent the ground truth goal positions. (a) Climbing stair. (b) Standing still. }
	\label{fig3}
\end{figure}

\begin{figure}[h!]
	\centering
	\begin{tabular}{cc}
		\includegraphics[width=0.45\textwidth]{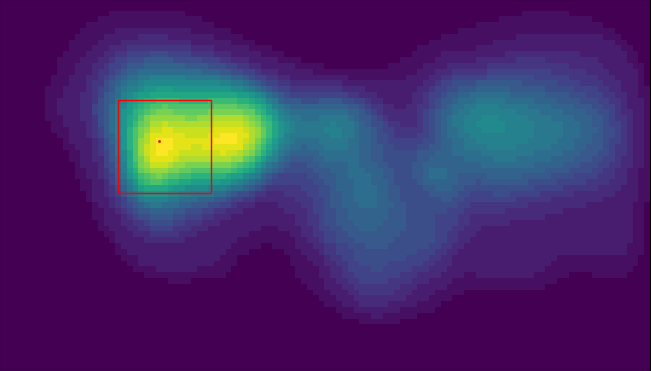}&
		\includegraphics[width=0.45\textwidth]{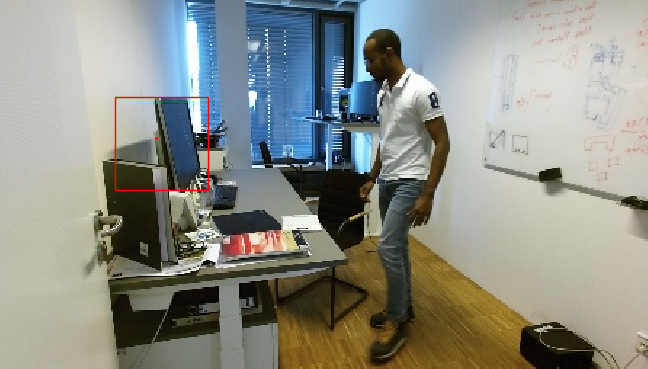}\\
		\includegraphics[width=0.45\textwidth]{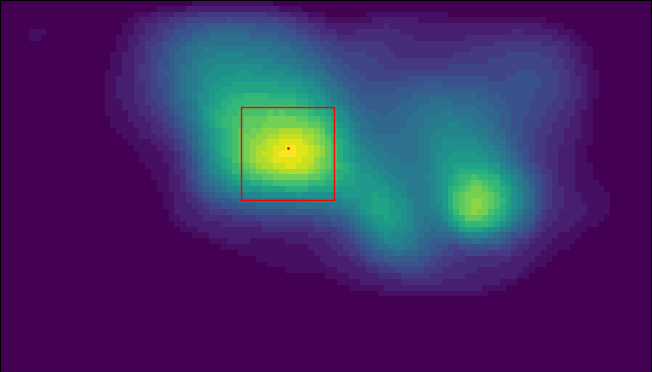}&
		\includegraphics[width=0.45\textwidth]{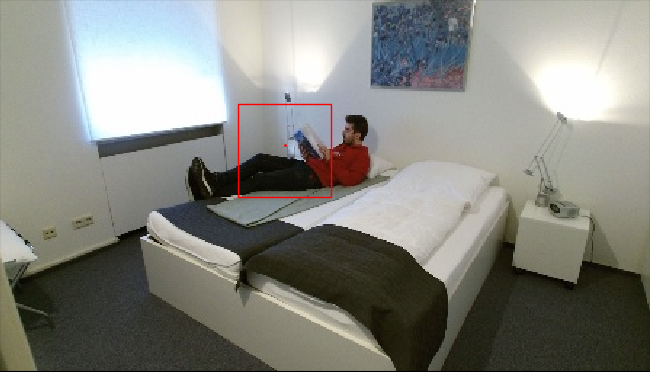}\\
	\end{tabular}
	\caption{\textbf{Deterministic Predictions in PROX.} The left column displays the predicted goal heatmaps, from which the goal areas are extracted. The right column presents the corresponding scenes with the predicted goals.}
	\label{fig6}
\end{figure}

\paragraph{Deterministic Predictions.} Our CVAE also shows promising results for deterministic target prediction without temperature scaling. Figure \ref{fig3} (a) and (b) present single forecasts in scenes where only one fixed behaviour can be performed. The prediction errors of these two scenes are 5 and 2 pixels respectively. Figure \ref{fig6} shows goal heatmaps of deterministic predictions and the corresponding scenes in the real world. One subject is going to the table and the other continues to lie on the bed.

\begin{figure}
	\begin{tabular}{ccc}
		\includegraphics[width=0.33\textwidth]{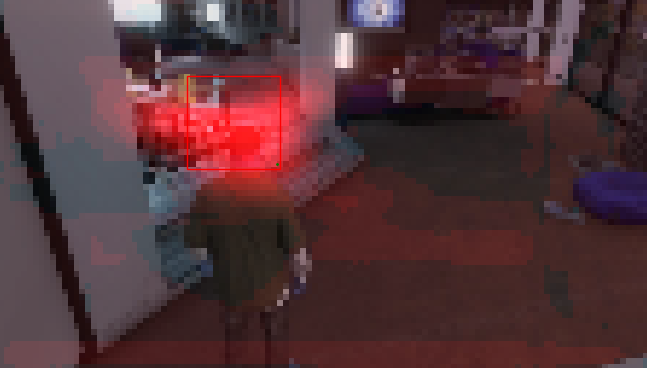}&
		\includegraphics[width=0.33\textwidth]{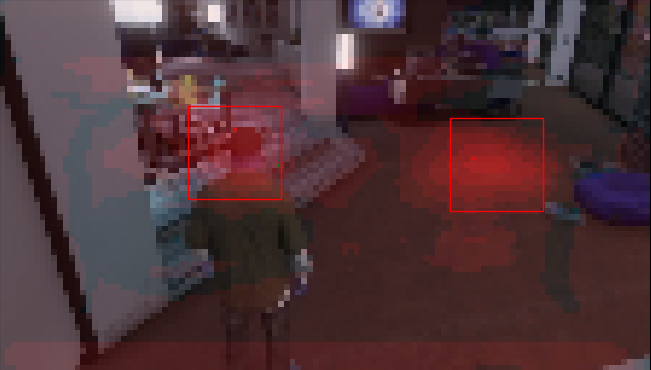}&
		\includegraphics[width=0.33\textwidth]{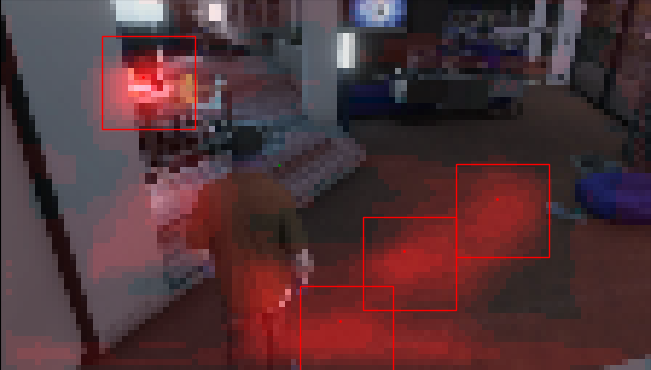}\\
		
		\includegraphics[width=0.33\textwidth]{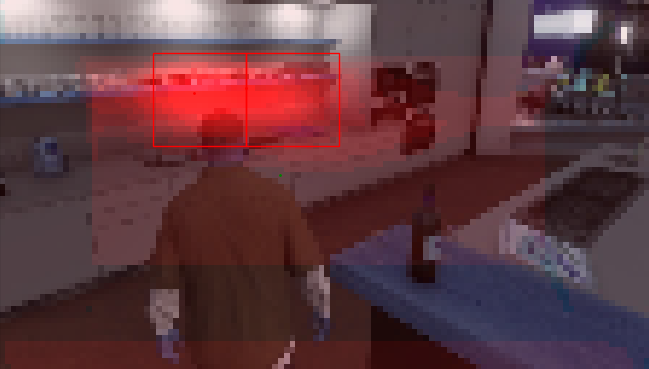}&
		\includegraphics[width=0.33\textwidth]{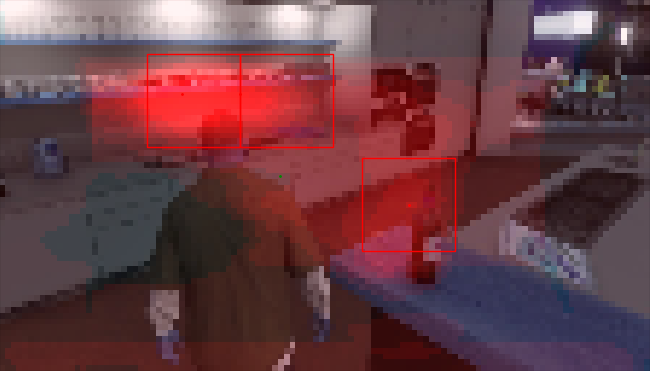}&
		\includegraphics[width=0.33\textwidth]{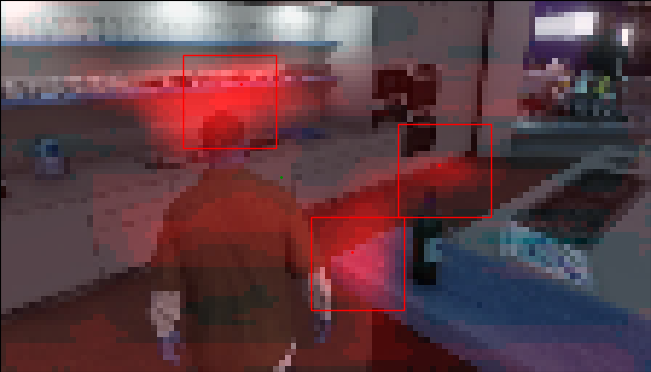}\\
	\end{tabular}
	\caption{\textbf{Probabilistic Predictions in GTA-IM.} From left to right by column, the temperature $\tau = 1, 200, 500$. $\tau = 1$ means deterministic prediction. }
	\label{fig4}
\end{figure}

\begin{figure*}
	\centering
	\caption{\textbf{Probabilistic Predictions in PROX.} From top to bottom by row, the temperature $\tau = 100, 200, 300$. The subject would turn left and walk through the open space to the door, or back to the table behind. }
	\begin{tabular}{ccc}
		\includegraphics[width=0.33\textwidth]{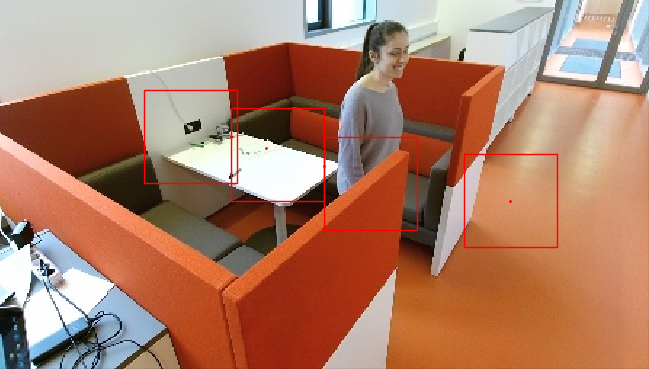}&
		\includegraphics[width=0.33\textwidth]{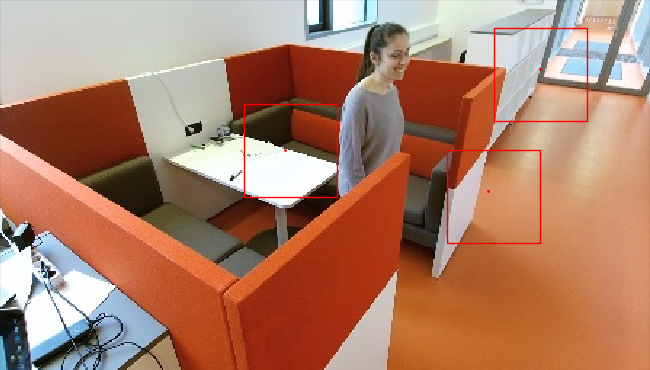}&
		\includegraphics[width=0.33\textwidth]{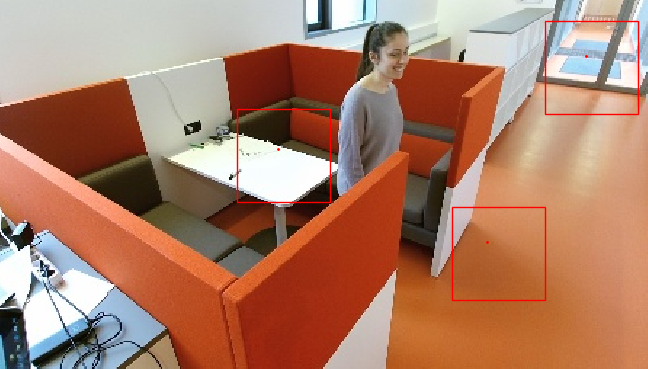}\\
		\includegraphics[width=0.33\textwidth]{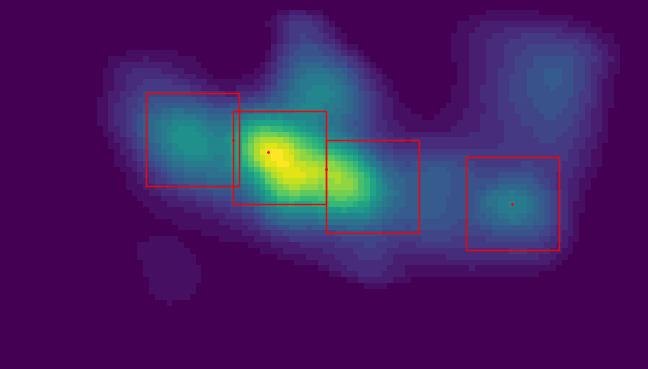}&
		\includegraphics[width=0.33\textwidth]{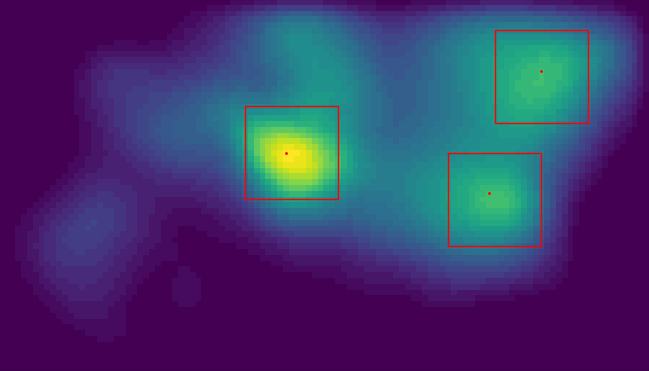}&
		\includegraphics[width=0.33\textwidth]{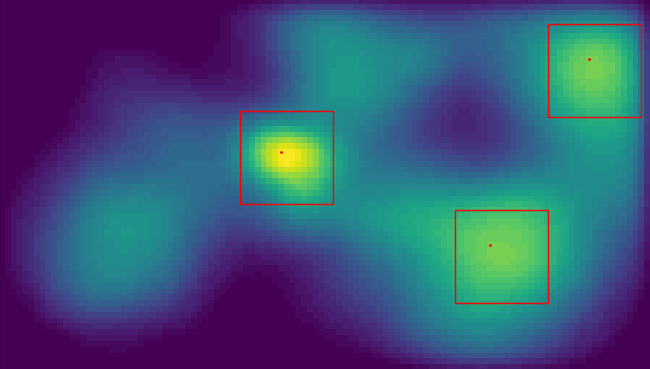}\\
	\end{tabular}
	
	\label{fig8}
\end{figure*}

\paragraph{Probabilistic Predictions.} When the temperature $\tau > 1$, the model turns to generative prediction. As shown in Figure \ref{fig4}, there is a trend towards more diversity as $\tau$ increases, with more forecasts being delivered. Figure \ref{fig8} presents various predictions with increasing temperature $\tau$ in the real world. A larger $\tau$ covers more possibilities of potential goals within the scene.

\begin{figure*}
	\centering
	\caption{\textbf{Generalization across Scenarios in GTA-IM.} (a) Generalisation to interactive objects. (b) Generalisation to unseen scenes. }
	\label{fig5}
	\begin{tabular}{cc}
		\includegraphics[width=0.45\textwidth]{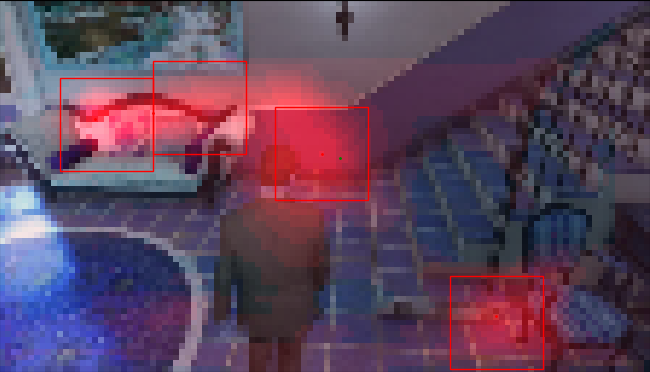}&
		\includegraphics[width=0.45\textwidth]{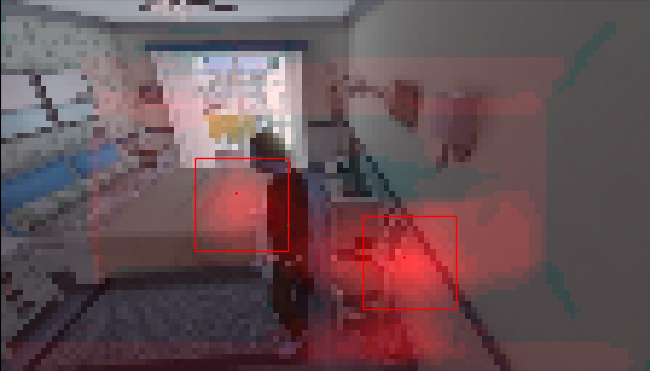}\\
		(a)&(b)
	\end{tabular}
\end{figure*}

\begin{figure*}
	\centering
	\caption{\textbf{Generalization across Scenarios in PROX.} Predictions in scenarios not seen during training. (Top) The subject walks through the open space. (Bottom) The subject sits on the chair, or may go to the bed.}
	\label{fig7}
	\begin{tabular}{cc}
		\includegraphics[width=0.45\textwidth]{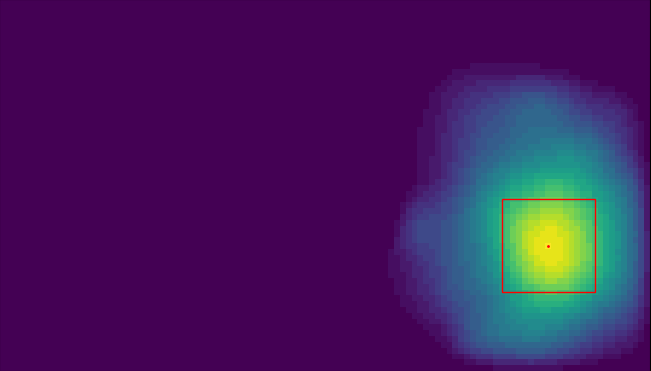}&
		\includegraphics[width=0.45\textwidth]{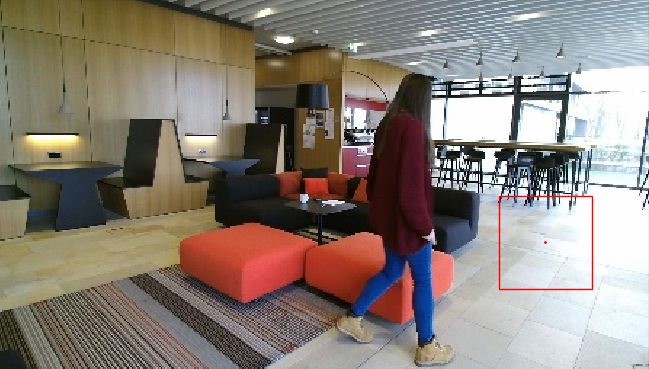}\\
		\includegraphics[width=0.45\textwidth]{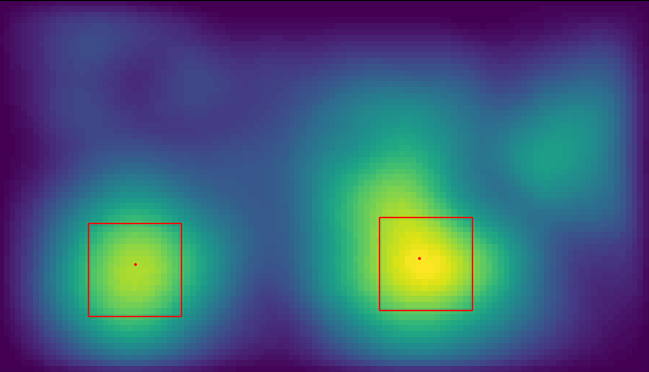}&
		\includegraphics[width=0.45\textwidth]{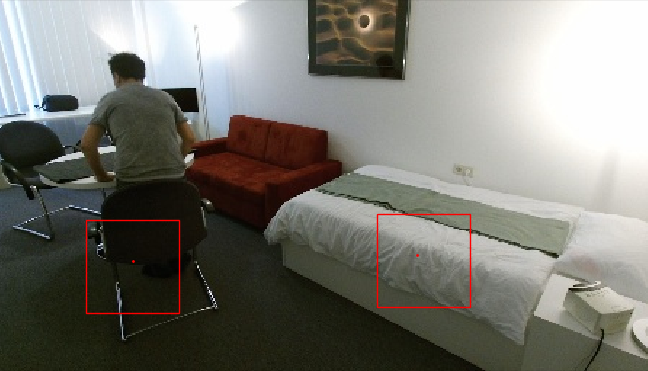}\\
	\end{tabular}
	
\end{figure*}

\paragraph{Generalization across Scenarios.}
Although our model is trained on real movement sequences of subjects without diverse manual labels, it shows generalization to behavioural options that never happened and to scenes unseen during training. Figure \ref{fig5} (a) anticipates the sofa as one of the potential target areas, even though the subjects never go to this sofa in our training data. Figure \ref{fig5} (b) is a bedroom not encompassed within our training dataset, but our model still delivers highly plausible forecasts. In this moment, the subject could either go to the bed or sit on the chair positioned behind. Two scenes of PROX dataset that are not seen during training, as shown in Figure \ref{fig7}, also yield promising predictions. In particular, the bedroom depicted at the bottom right of Figure \ref{fig7} has a relatively compact layout.

\begin{table}[h!]
	\caption{The final displacement errors (FDEs) in pixel of our model compared to the baseline \cite{longterm} in GTA-IM dataset and PROX dataset. The lower, the better. We set the temperature $\tau$ to 100 in the evaluation of probabilistic predictions.}\label{gta} \centering
	\begin{tabular}{|l|l|l|l|l|l|l|l|l|}
		\hline
		\multirow{1}{*}{} & \multicolumn{4}{c|}{GTA-IM}& \multicolumn{4}{c|}{PROX} \\
		\cline{2-5} \cline{6-9}
		Method & $F = 1$ & $F = 3$ & $F = 5$ & $F = 10$ & $F = 1$ & $F = 3$ & $F = 5$ & $F = 10$ \\
		\hline
		GoalNet (min) & 22.5 & 23.4 & 22.7 & 17.1 & 24.4 & 27.0 & 24.9 & 21.5\\
		Ours (min) & \textbf{10.9} & \textbf{7.9} & \textbf{6.8} & \textbf{5.2} & \textbf{10.38} & \textbf{7.0} & \textbf{6.2} & \textbf{5.4}\\
		GoalNet (avg) & $-$ & 38.2 & 37.9 & 38.3& $-$ & 40.1 & 39.9 & 39.5\\
		Ours (avg) & $-$  & \textbf{13.1} & \textbf{13.3} & \textbf{12.9}& $-$  & \textbf{10.1} & \textbf{10.1} & \textbf{10.0}\\
		\hline
	\end{tabular}
\end{table}

\subsubsection{Quantitative Analysis}
Average final displacement errors ($FDEs_{avg}$) and minimal final displacement errors ($FDEs_{min}$) are analyzed as defined in Equation \ref{metric} and \ref{metric_min}.
Table\ref{gta} lists the FDEs of deterministic prediction ($F = 1$) and probabilistic prediction with varying numbers of samples ($F = 3, 5, 10$) for our method and the baseline in GTA-IM and PROX dataset respectively. Overall, our minimal and average errors in both probabilistic and deterministic forecasting are significantly lower than the baseline in both benchmarking datasets. Our $FDEs_{min}$ continue to decrease as the number of forecasts increases, indicating that more accurate predictions are generated. Our $FDEs_{avg}$ remain nearly constant even with the increasing number of forecasts, showing that the generation quality is stable, as the temperature parameter $\tau$ remains constant during this evaluation.

Moreover, in order to enable real-time deployment of our model on edge devices for robotic applications, the Average Inference Time (AIT) was also measured. Our average inference time of 43.8ms guarantees real-time capability for applications, and is also around 11\% faster than the baseline, as listed in Table \ref{ait}.

\begin{table}[h]
	\caption{Average Inference Time (AIT) in millisecond of our model and the baseline \cite{longterm}.}\label{ait} \centering
	\begin{tabular}{|l|l|l|}
		\hline
		Method &  GoalNet & Ours \\
		\hline
		Average Inference Time (AIT) & 49.4 
		 & \textbf{43.8}\\
		\hline
	\end{tabular}
\end{table}


\subsubsection{Ablation Study}

We conducted ablation studies to test the effects of the components in our method, as listed in Table \ref{tab_abla}. First, the dimension of the latent space was investigated. With a smaller dimension of latent space, the performance of deterministic and probabilistic predictions is worse, indicating that 30 dimensions are not sufficient to represent features. 

\begin{table*}[h]
	\caption{Ablation Studies. \XSolidBrush denotes the impossibility of generating predictions. $F = 3$ in the evaluation of probabilistic prediction.}\label{tab_abla} \centering
	\begin{tabular}{|l|l|l|l|l|}
		\hline
		\multirow{2}{*}{Method} & \multicolumn{2}{c|}{ Deterministic Prediction} & \multicolumn{2}{c|}{Probabilistic Prediction} \\
		\cline{2-5}
		&$FDE_{min}$ (GTA) & $FDE_{min}$ (PROX) & $FDE_{min}$ (GTA) & $FDE_{min}$ (PROX) \\
		\hline
		Ours (Z = 30) & 15.6 & 14.9 & 12.9 & 12.3 \\
		\hline
		Ours (condition on history trajectory) & 18.3 & 15.5 & \XSolidBrush& \XSolidBrush\\
		Ours (condition on torso heatmap) & 13.3 & 12.4 & \XSolidBrush& \XSolidBrush\\
		w/o condition & \XSolidBrush & \XSolidBrush & \XSolidBrush& \XSolidBrush \\
		\hline
		w/o multi-maxima & 20.4 & 21.3 & 17.7 & 19.1\\
		\hline
		Our full model & 10.9 & 10.3 & 7.9 & 7.0 \\
		\hline
	\end{tabular}
\end{table*}

Second, we tested different conditions of the CVAE. Our model, whether conditioned on the human history trajectory, a torso heatmap, or left unconditioned, is not capable of probabilistic prediction, because it cannot generate multiple maxima. In particular, when conditioning on a set of human trajectory points, it is easy to overfit to positions close to the last trajectory point in the history, indicating that the scene context does not contribute to the prediction. Its deterministic forecasting is also less accurate. Conditioning on a torso heatmap means $H_{J, 0}$ only contains one Gaussian around the torso position without other joints to represent the human position. It performs slightly worse than our proposed model in deterministic prediction, while it also has no capability for probabilistic prediction because it can only generate a single maximum. Additionally, we removed the condition in our training and inference, which makes our model a VAE. The input modality becomes a scene image without explicit information of the human. It is found out that no scene structure is learned and no target can be predicted. This proves the necessity of human information, as our model is not designed to detect humans within the scene image and intuitively the current location is required to derive the future destination. 

Third, extracting only one maximum from the predicted heatmap has larger errors than extracting multiple maxima, which demonstrate the positive effect of multi-maxima that takes local maxima into account.

The temperature parameter $\tau$ introduces variation into sampling process of the latent space \cite{generativebook1st}. The setting of this parameter is a trade-off between anticipating a single but possibly more precise destination and generating more flexible predictions according to the scene. Table \ref{tab_tau} investigates the influence of $\tau$ in probabilistic forecasting. As the $\tau$ rises, the average FDE continues to increase. Because a higher temperature scaling factor yields more bad samples due to more variation. The averaged minimum FDE is relatively stable up to $\tau = 300$, which demonstrates the stable capability to generate good samples.

\begin{table}[h]
	\caption{Ablation study on the parameter $\tau$ for inference via generation. $F = 3$.}\label{tab_tau} \centering
	\begin{tabular}{|l|l|l|l|l|l|}
		\hline
		Parameters&$\tau = 50$&$\tau = 80$&$\tau = 200$&$\tau = 300$&$\tau = 500$\\
		\hline
		$FDE_{avg}$&11.9&12.1&15.1&16.1&19.8\\
		$FDE_{min}$&7.6&8.6&8.6&8.5&9.9\\
		\hline
	\end{tabular}
\end{table}

\subsubsection{Limitations and Future Work}
Our evaluations show that our multi-goal prediction provides more accurate samples than just predicting one deterministic goal. But so far the generation by random sampling is not fully controlled to only sample in the real data distribution of the latent space. Currently, we are mostly able to distinguish good samples from bad ones, as the bad samples usually have a lower variance of values or lower maximal values, indicating low confidence in the prediction. The efficient generation of effective samples is still a challenging problem \cite{dlow} and is planned to be explored in the future.

In the robotic applications, apart from selecting a final prediction with the highest confidence score among the $M$ forecasts, the historical trajectory of the human can also be used to guide the selection of an optimal prediction. This is because a future position can be estimated based on the current position and velocity. Therefore, the final prediction could be chosen by selecting a forecast that minimizes the distance to the simple estimate of the future position. The prediction accuracy using confidence scores or using historical trajectory could be compared in the future.

\section{\uppercase{Conclusion}}
In summary, our work presents a scene-aware method for anticipating diverse human movement goals. Firstly, we trained the CVAE model to deliver pixel-wise outputs, and then in post-processing extracted target areas represented by bounding boxes and predicted the coordinates of the human movement goal. Secondly, in the inference stage, diverse goals are obtained by sampling in the latent space via temperature scaling. Besides, our model shows the generalization capability across interactive objects in the scene and unknown scenarios, since it has been trained to implicitly understand the features of  scenes within the target area. The evaluation shows that our method significantly outperforms the baseline. In the future, the integration of this method into a collaborative robot environment is planned, for instance, our predicted goal heatmap could be transferred to the social cost layer for robot navigation.\\

\begin{credits}
\subsubsection{\ackname} This work was funded by Deutsche Forschungsgemeinschaft (DFG,
German Research Foundation) under grant number 502483052.

\subsubsection{\discintname}
The authors have no competing interests to declare that are relevant to the content of this article. 
\end{credits}
%
%
%
\bibliographystyle{splncs04}
%

\end{document}